\documentclass{article}
\usepackage{racer,times}
\usepackage{amsmath,amssymb}
\usepackage{graphicx}
\usepackage{booktabs}
\usepackage{multirow}
\usepackage[table]{xcolor}
\usepackage{enumitem}
\usepackage{subcaption}
\usepackage{url}
\usepackage{hyperref}

\title{RACER: Reflective Agent Coupling Query\\
Interpretation and Tool-Based Retrieval for\\
Frame Selection in Long Video Understanding}
\author{Yiyang Huang$^{1}$ Yitian Zhang$^{1}$ Yizhou Wang$^{1}$ Jianglin Lu$^{1}$ Qihua Dong$^{1}$\\
\textbf{Hailing Wang$^{1}$ Huimin Zeng$^{1}$ Mingyuan Zhang$^{1}$ Yun Fu$^{1,2}$}\\
$^{1}$Department of Electrical and Computer Engineering, Northeastern University\\
$^{2}$Khoury College of Computer Science, Northeastern University\\
\texttt{\{huang.yiyan, y.fu\}@northeastern.edu}\\
}

\hypersetup{
    pdftitle={RACER: Reflective Agent Coupling Query Interpretation and Tool-Based Retrieval for Frame Selection in Long Video Understanding},
    pdfauthor={Yiyang Huang, Yitian Zhang, Yizhou Wang, Jianglin Lu, Qihua Dong, Hailing Wang, Huimin Zeng, Mingyuan Zhang, Yun Fu},
    pdfsubject={Long-video frame selection},
    pdfkeywords={RACER, video understanding, frame selection, reflective agents},
    colorlinks=false,
    pdfborder={0 0 1},
    linkbordercolor={1 0 0},
    citebordercolor={0 1 0},
    urlbordercolor={0 1 1}
}

\begin{document}
\maketitle

\begin{abstract}
Video large language models (Vid-LLMs) excel at diverse video-language tasks by reasoning over selected frames. However, frame selection for long videos remains challenging, as it requires retrieving relevant frames distributed across segments from a large candidate pool given complex queries.
This paper investigates dominant approaches to long-video frame selection from a task-decomposition perspective, identifying two key challenges: the Query Comprehension Gap in similarity-based methods and the Interpretation--Selection Gap in judgment-based methods.
To address them, we propose RACER, a training-free reflective agentic framework that decomposes long-video frame selection into query interpretation driven by a lightweight Vid-LLM and evidence localization supported by an embedding model serving as a retrieval tool.
Specifically, the Vid-LLM is responsible solely for reformulating the complex query into sub-queries that make implicit information requirements explicit, mitigating the Query Comprehension Gap. Meanwhile, the retrieval tool leverages these sub-queries to localize relevant evidence, relieving the Vid-LLM of direct frame selection and thus addressing the Interpretation--Selection Gap. Finally, the retrieved frames are fed back to the Vid-LLM for sub-query refinement, forming a reflection loop that iteratively improves query interpretation and frame selection.
Experiments across multiple benchmarks show that RACER consistently improves long video understanding. Notably, RACER achieves effective frame selection even with limited-capability components, demonstrating that agentic integration enables these components to enhance more capable Vid-LLMs.
\textit{Code will be released.}
\end{abstract}

\section{Introduction}
\label{sec:intro}

Video large language models (Vid-LLMs)~\citep{videollama,videochatgpt,videollava,InternVideo3} integrate selected video frames with textual queries, achieving significant advances across diverse video-language tasks. However, key-frame selection in long videos remains challenging, as query-relevant evidence is scattered across segments within a vast candidate frame pool, while complex queries often fail to specify the required visual evidence. These challenges pose reliability and safety risks in autonomous driving~\citep{dolphins}, robotics~\citep{seedo}, and security~\citep{vad_vllm}.

Various approaches have been proposed for long-video frame selection, broadly following similarity-based and judgment-based paradigms.
Similarity-based methods use embedding models (e.g., CLIP~\citep{clip}) to score query-frame similarity for frame selection. BOLT~\citep{bolt} improves selection through inverse transform sampling, while Q-Frame~\citep{qframe} combines Gumbel-Max sampling with multi-resolution adaptation. AKS~\citep{aks} incorporates temporal coverage into selection, whereas MDP$^3$~\citep{mdp3} considers diversity and sequentiality in frame-set construction.
In contrast, judgment-based methods employ VLM/LLM decisions on candidate frames and segments, either assigning relevance levels or performing binary selection. Existing approaches include LLM assessment of VLM-generated captions~\citep{videotree,llovi}, iterative LLM selection of temporal search regions~\citep{videoagent}, and iterative VLM assignment of relevance levels to promising candidates~\citep{air}.
However, existing methods treat frame selection as a single task driven by similarity scoring or VLM/LLM judgment, leaving the processes of determining what evidence is needed and locating where it appears largely coupled.

\begin{figure}[t]
\begin{center}
\includegraphics[width=1.0\textwidth]{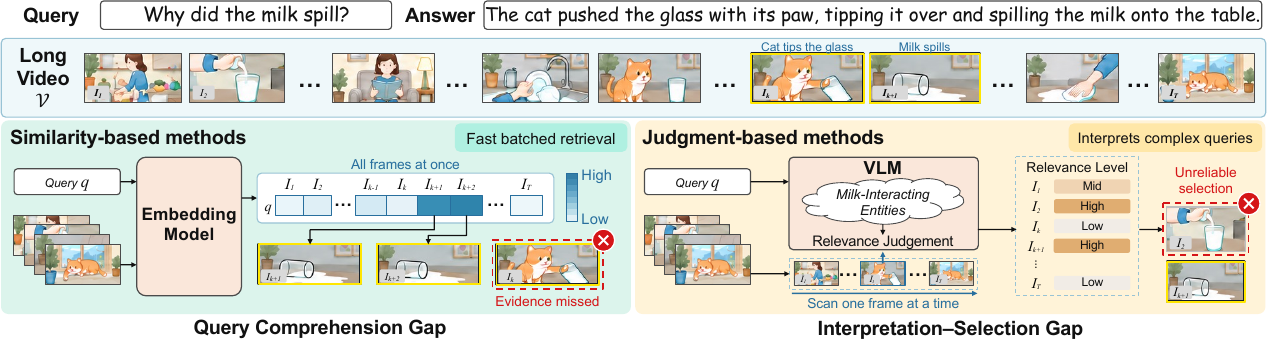}
\end{center}
\caption{
\textbf{Two key challenges in long-video frame selection.} Answering ``Why did the milk spill?'' requires evidence of the cat tipping the glass and the resulting spill. \textbf{Query Comprehension Gap (left):} similarity-based methods overlook implicit causal requirements, retrieving spilled-milk frames while missing the cat tipping the glass. \textbf{Interpretation--Selection Gap (right):} judgment-based methods understand implicit evidence requirements but fail to perform selection reliably, choosing a milk-pouring frame instead of the causal action.}
\label{fig:challenges}

\end{figure}

This paper investigates long-video frame selection from a task-decomposition perspective, identifying the \textbf{Query Comprehension Gap} in similarity-based methods when interpreting evidence requirements and the \textbf{Interpretation--Selection Gap} in judgment-based methods when localizing relevant evidence, as illustrated in Figure~\ref{fig:challenges}.
Similarity-based methods enable efficient and precise retrieval from large candidate pools, but embedding models may fail to capture the temporal, causal, and implicit information requirements of complex queries. Judgment-based methods can interpret complex queries and describe potentially relevant visual evidence, yet such understanding does not necessarily translate into reliable frame or segment selection.

To bridge these gaps, we propose \textbf{RACER}, a training-free reflective agentic framework that decomposes long-video frame selection into query interpretation assigned exclusively to a lightweight Vid-LLM and evidence localization delegated to an embedding model serving as a retrieval tool.
Specifically, the Vid-LLM observes a video overview and reformulates the complex query into sub-queries that make distinct implicit information requirements explicit to the embedding model, thereby mitigating the Query Comprehension Gap. Meanwhile, the retrieval tool grounds the original query and the generated sub-queries in candidate frames through similarity scoring, relieving the Vid-LLM of direct frame selection and addressing the Interpretation--Selection Gap. Finally, the retrieved frames are fed back to the Vid-LLM for sub-query refinement, forming a reflection loop that iteratively improves query interpretation and frame selection.

Experiments demonstrate that RACER consistently improves performance on long-video understanding benchmarks, including Video-MME~\citep{videomme}, MLVU~\citep{mlvu}, and LongVideoBench~\citep{longvideobench}. These gains extend across multiple Vid-LLMs, including LLaVA-OneVision~\citep{llavaov}, LLaVA-Video~\citep{llavavideo}, Qwen2.5-VL~\citep{qwen25vl}, and Qwen3-VL~\citep{qwen3vl}. Notably, RACER achieves effective key-frame selection even with limited-capability components, demonstrating that agentic integration enables these components to enhance more capable Vid-LLMs.

Our contributions are summarized as follows:
\begin{itemize}[noitemsep, topsep=0pt, parsep=0pt, partopsep=0pt, leftmargin=*]
\item We investigate long-video frame selection from a task-decomposition perspective, identifying the \textbf{Query Comprehension Gap} in similarity-based methods and the \textbf{Interpretation--Selection Gap} in judgment-based methods.

\item We propose \textbf{RACER}, a training-free reflective agentic framework that decomposes long-video frame selection into query interpretation assigned exclusively to a lightweight Vid-LLM and evidence localization delegated to an embedding model serving as a retrieval tool.

\item Comprehensive experiments demonstrate RACER's effectiveness across multiple long-video understanding benchmarks and Vid-LLM families. The results further show that agentic integration enables limited-capability components to enhance more capable Vid-LLMs.
\end{itemize}

\section{Related Work}
\label{sec:related}

\subsection{Video Large Language Models (Vid-LLMs)}
Early Vid-LLMs connect video features to LLMs through query-based adapters~\citep{videochat,videollama} or spatiotemporal pooling~\citep{videochatgpt}. Video-LLaVA~\citep{videollava} aligns image and video representations, while VILA~\citep{vila} adopts interleaved pretraining. More recent models~\citep{llavaov,llavavideo,qwen25vl,qwen3vl,internvl3}, further improve unified image-video modeling, instruction tuning, and spatiotemporal representation.
For long videos, MovieChat~\citep{moviechat} and LLaMA-VID~\citep{llamavid} compress visual inputs through memory or compact tokens, while LongVA~\citep{longva} and LongVU~\citep{longvu} extend context capacity through adaptive spatiotemporal compression. Nevertheless, Vid-LLMs process only limited sampled frames, making effective key-frame selection critical for video understanding.

\subsection{Key-Frame Selection for Vid-LLMs}
Existing Vid-LLM frame-selection methods broadly follow similarity-based and judgment-based paradigms. Similarity-based methods use embedding models, such as CLIP~\citep{clip}, to score query--frame similarity for frame selection. Building on these similarity scores, BOLT~\citep{bolt} uses inverse transform sampling, while Q-Frame~\citep{qframe} combines Gumbel-Max sampling with multi-resolution adaptation. AKS~\citep{aks} adds temporal coverage, whereas MDP$^3$~\citep{mdp3} considers diversity and sequentiality. Other methods adapt frame allocation using temporal structure or similarity distributions~\citep{crest,adaq}.
In contrast, judgment-based methods use VLM/LLM decisions to assign relevance levels or select candidate frames and temporal regions. Caption-mediated judgments support hierarchical frame selection~\citep{videotree} and supporting-frame identification in grounded VideoQA~\citep{llovi}. VideoAgent~\citep{videoagent} iteratively selects temporal search regions for progressive video analysis, while A.I.R.~\citep{air} assigns relevance levels to promising candidates and expands the search around validated frames. Other approaches learn selection decisions through localization~\citep{sevila}, rankings derived from downstream prediction losses~\citep{framevoyager}, or autoregressive policies trained with answer feedback~\citep{refocus}.
While effective, these methods primarily treat frame selection as a unified task driven by similarity scoring or VLM/LLM judgment, leaving evidence requirement interpretation and localization largely coupled, a limitation this work aims to address through task decomposition.

\section{Method}
\label{sec:method}

\subsection{Challenges in Dominant Frame Selection Approaches}
\label{sec:challenge_analysis}
Effective frame selection requires interpreting query-implied evidence requirements and reliably localizing corresponding frames within a large candidate pool. However, similarity-based methods face the Query Comprehension Gap in interpreting complex queries, whereas judgment-based methods face the Interpretation--Selection Gap in selecting relevant frames. This section analyzes these two challenges.

\begin{figure*}[t]
\centering
\begin{subfigure}[t]{0.49\textwidth}
    \centering
    \includegraphics[width=0.49\linewidth]{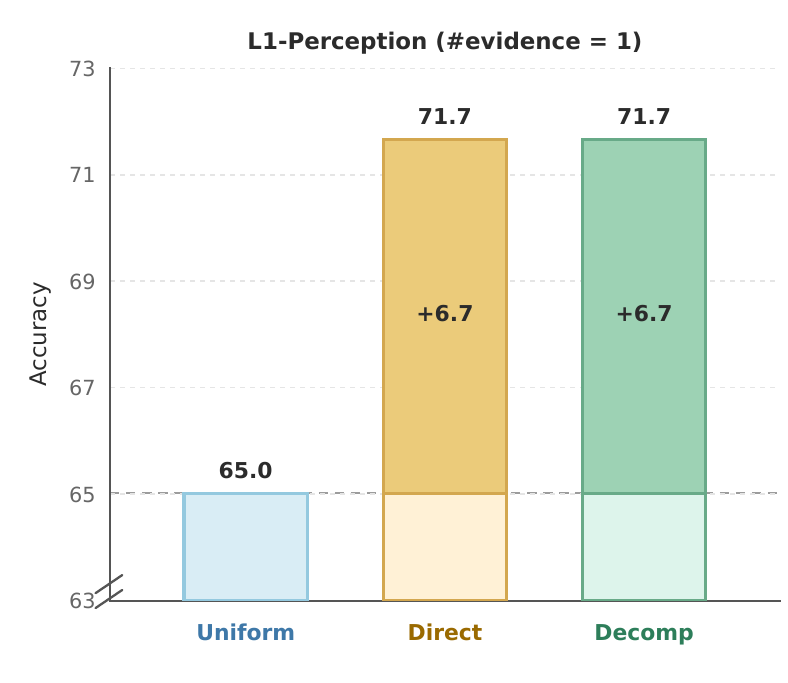}
    \hfill
    \includegraphics[width=0.49\linewidth]{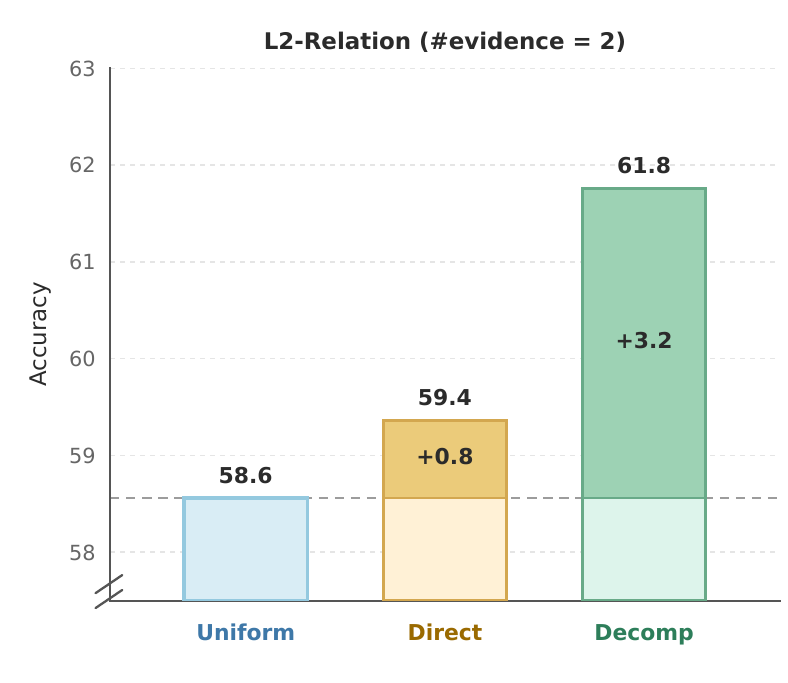}
    \caption{\textbf{Query Comprehension Gap.}
    Accuracy of Uniform Sampling, Direct (CLIP selection with the original query), and Decomp (CLIP selection with the original query and VLM-generated sub-queries) on single-evidence (left) and multi-evidence (right) queries.}
    \label{fig:challenge1}
\end{subfigure}
\hfill
\begin{subfigure}[t]{0.49\textwidth}
    \centering
    \includegraphics[width=0.49\linewidth]{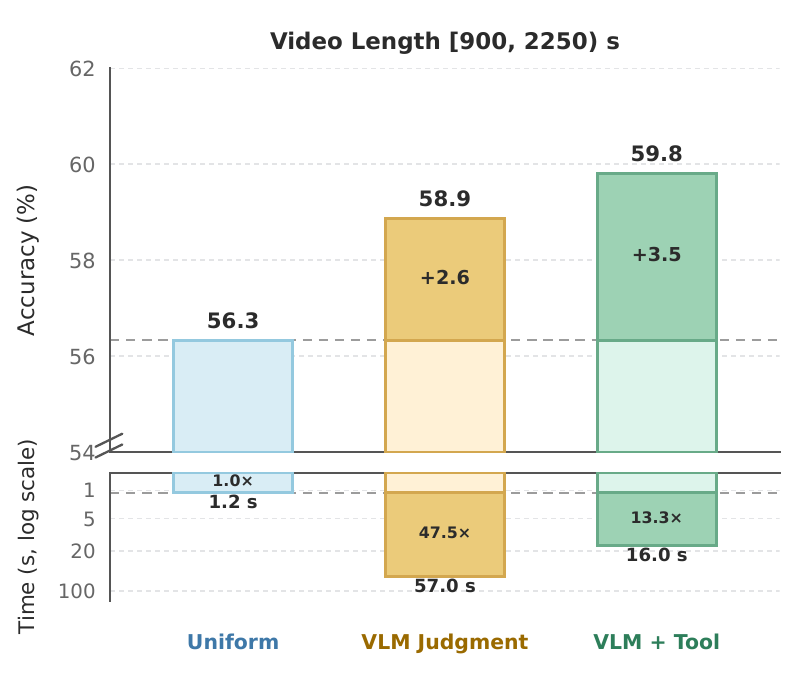}
    \hfill
    \includegraphics[width=0.49\linewidth]{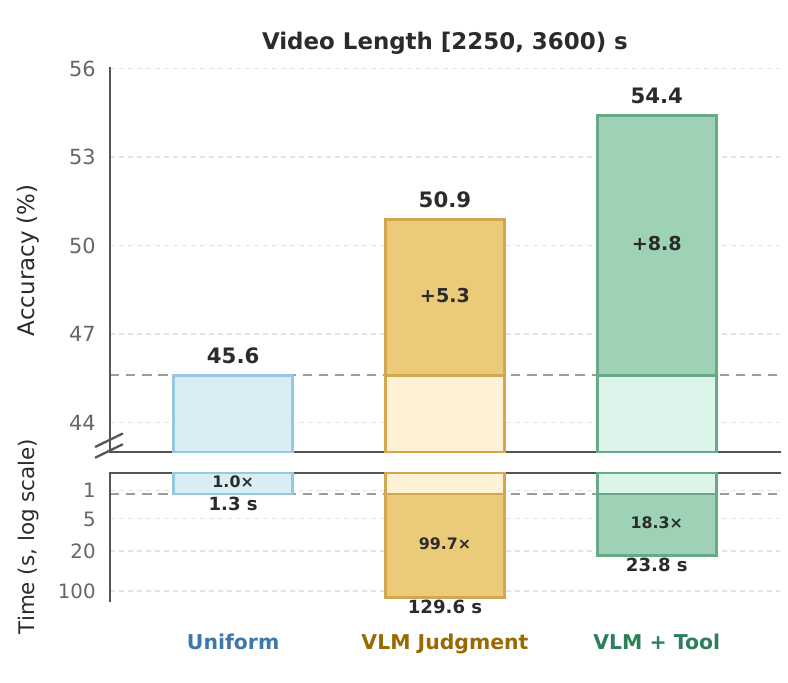}
    \caption{\textbf{Interpretation--Selection Gap.}
    Accuracy (top) and time cost (bottom) of Uniform Sampling, VLM Judgment (VLM-assigned frame relevance levels), and VLM+Tool (VLM query decomposition with CLIP frame selection) across two video-duration groups.}
    \label{fig:challenge2}
\end{subfigure}

\caption{
\textbf{Empirical analysis of two gaps on LongVideoBench subsets.}
\textbf{(a)} Embedding models such as CLIP handle single-evidence queries effectively but struggle with implicit requirements in complex queries, which query decomposition helps make explicit.
\textbf{(b)} VLMs are more effective at interpreting complex queries and describing relevant evidence than at directly selecting frames, as using these interpretations to guide tool-based selection yields higher accuracy and lower latency than direct VLM selection, especially for long videos.
}
\label{fig:challenge_analysis}
\end{figure*}

\subsubsection{Query Comprehension Gap in Similarity-Based Methods}
\label{sec:challenge_comprehension}

Similarity-based methods select frames by query--frame embedding similarity. While effective for simple queries requiring a single salient evidence target, they are less effective for complex queries requiring multiple evidence targets linked by temporal, causal, or relational dependencies.

To reveal this challenge, two LongVideoBench~\citep{longvideobench} subsets differing in query complexity are considered: single-evidence L1-Perception and multi-evidence L2-Relation. Three methods are compared: uniform sampling, direct selection using only the original query, and decomposed selection using both the original query and sub-queries generated through query decomposition by a lightweight Vid-LLM (Qwen2.5-VL-3B) observing 32 uniformly sampled overview frames. Both selection methods select the top-32 frames by CLIP query--frame similarity. All methods use 32 frames for VideoQA with LLaVA-OneVision.
As shown in Figure~\ref{fig:challenge1}, direct and decomposed selection yield equally substantial accuracy gains over uniform sampling on single-evidence L1-Perception queries, whereas on multi-evidence L2-Relation queries, the gain is only 0.8 for direct selection compared with 3.2 for decomposed selection. These results illustrate the \textbf{Query Comprehension Gap} in similarity-based methods: embedding models struggle to capture implicit requirements in complex queries, while query decomposition makes these requirements explicit.

\subsubsection{Interpretation--Selection Gap in Judgment-Based Methods}
\label{sec:challenge_selection}

Judgment-based methods employ VLM/LLM decisions on candidate frames and segments. While effective at understanding complex queries and describing potentially relevant evidence, these models can be inefficient and unreliable when assigning relevance levels or making binary selections over large candidate pools.

To reveal this challenge, the subset of LongVideoBench~\citep{longvideobench} with the widest duration range is considered and divided by video duration into two groups: $900$--$2250$ and $2250$--$3600$ seconds. Three methods are compared: uniform sampling, VLM judgment-based selection assigning a relevance level to each frame, and tool-augmented VLM selection, where the VLM only decomposes the query into sub-queries based on 32 uniformly sampled overview frames, while CLIP serves as a retrieval tool to score frames against the original query and sub-queries. Both VLM variants use Qwen2.5-VL-3B and select the top-32 frames by relevance levels or similarity scores. All methods use 32 frames for VideoQA with LLaVA-OneVision.
As shown in Figure~\ref{fig:challenge2}, both VLM variants improve accuracy over uniform sampling, while tool-augmented VLM selection consistently outperforms VLM judgment-based selection in accuracy and efficiency, achieving an additional accuracy gain of 3.5 and a $5.4\times$ speedup in the longer-duration group. These results illustrate the \textbf{Interpretation--Selection Gap} in judgment-based methods: VLM/LLM judgment does not ensure reliable and efficient frame selection. The higher accuracy and efficiency of retrieval guided by VLM-generated interpretations suggest that VLMs are better suited to interpreting evidence requirements than selecting frames, especially for longer videos.

\begin{figure*}[t]
    \centering
    \includegraphics[width=\textwidth]{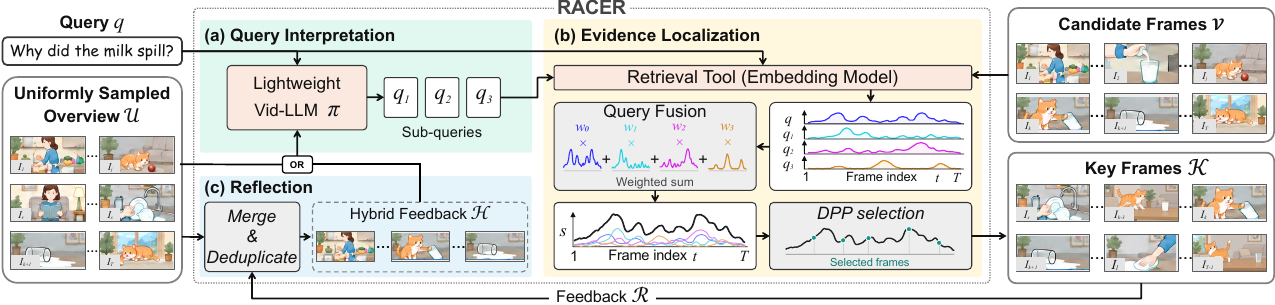}
    \vspace{-8pt}
\caption{
\textbf{Overview of RACER.}
RACER begins with \textbf{(a) Query Interpretation}, where a lightweight Vid-LLM uses a uniformly sampled overview to decompose the query into complementary sub-queries, making implicit evidence requirements explicit.
Then, \textbf{(b) Evidence Localization} employs an embedding-based retrieval tool to score candidate frames against the original query and sub-queries, followed by entropy-adaptive score fusion and DPP selection to obtain relevant and visually diverse key frames.
Finally, \textbf{(c) Reflection} merges feedback frames with the overview, removing duplicates and the least visually aligned overview frames to maintain the frame budget. The resulting hybrid overview guides sub-query refinement for the next selection round.
}
    \label{fig:pipeline}
\end{figure*}

\subsection{RACER: Reflective Agent Coupling Query Interpretation and Tool-Based Retrieval}

Building on these observations, we propose \textbf{RACER}, a training-free reflective agentic framework that decomposes long-video frame selection into query interpretation and evidence localization, addressing the Query Comprehension Gap and the Interpretation--Selection Gap, respectively.
As illustrated in Figure~\ref{fig:pipeline}, RACER begins with query interpretation: a lightweight Vid-LLM observes a video overview and decomposes the query into sub-queries, making distinct implicit information requirements explicit to the embedding model. For evidence localization, the embedding model serves as a retrieval tool, grounding the original query and sub-queries in candidate frames through similarity scoring and relieving the Vid-LLM of direct frame selection. Finally, retrieved frames are fed back to the Vid-LLM for sub-query refinement, forming a reflection loop that iteratively improves query interpretation and frame selection.

\subsubsection{Task Formulation: Frame Selection in Long Video Understanding}
\label{sec:method_formulation}

Frame selection enables Vid-LLMs with limited frame budgets to understand long videos. Given candidate frames $\mathcal{V}=\{I_t\}_{t=1}^{T}$ and a budget $N\ll T$, a selector $\mathcal{S}$ selects key frames $\mathcal{K}$ based on a query $q$:
\begin{equation}
\mathcal{K}=\mathcal{S}(\mathcal{V},q),
\qquad
\mathcal{K}\subset\mathcal{V},\ |\mathcal{K}|=N.
\label{eq:selection}
\end{equation}
The selected frames $\mathcal{K}$ and query $q$ are then provided to a Vid-LLM $\mathcal{F}$ to generate an answer:
\begin{equation}
A=\mathcal{F}(\mathcal{K},q).
\label{eq:answer}
\end{equation}

\subsubsection{Query Interpretation}
\label{sec:method_comprehension}

As discussed in Section~\ref{sec:challenge_comprehension}, the Query Comprehension Gap limits embedding models in handling complex queries requiring interrelated evidence. To address this gap, the Query Interpretation module decomposes the original query into sub-queries, making implicit retrieval needs explicit.

Specifically, the module employs a lightweight Vid-LLM $\pi$ to generate sub-queries $\{q_1,\ldots,q_M\}$ from query $q$ and a video overview comprising $N$ frames uniformly sampled from the candidate pool $\mathcal{V}$:
\begin{equation}
\{q_1,\ldots,q_M\}=\pi\!\left(q,\mathrm{Uniform}(\mathcal{V},N)\right).
\label{eq:decompose}
\end{equation}
The decomposition is guided by the prompt in Table~\ref{tab:query_prompt}, which instructs $\pi$ to generate complementary, visually grounded sub-queries, each targeting a distinct information requirement implicit in the original query.

\begin{table*}[t]
\centering

\begin{minipage}[t]{0.47\textwidth}
\vspace{0pt}
\centering
\caption{Prompt for Query Interpretation.}
    \vspace{-8pt}
\label{tab:query_prompt}
\scriptsize

\begin{tabular}{@{}p{\linewidth}@{}}
\toprule
\textbf{Query Interpretation Prompt} \\
\midrule

You are analyzing a video question. To answer the main question well,
you sometimes need to first know the answers to simpler sub-questions
about the video. Watch the video and propose 2--3 such sub-questions.

\smallskip
\textbf{Each sub-question must:}
\begin{itemize}[leftmargin=*, noitemsep, topsep=0pt]
    \item Be answerable by looking at a single still frame---ask about
    visible objects, people, places, attributes, settings, or counts.
    Avoid asking about actions, sequences, order, changes over time,
    or ``what happens next''---those cannot be answered from one frame.

    \item Point to information the main question implicitly depends on,
    but does not itself ask about directly.

    \item Be phrased in plain natural language as a question rather
    than a statement or caption.

    \item Be different from the main question and from each other.
\end{itemize}

\smallskip
\textbf{Main question:} \texttt{\{query\}} \\
\bottomrule
\end{tabular}
\end{minipage}\hfill%
\begin{minipage}[t]{0.49\textwidth}
\vspace{0pt}
\centering
\caption{Prompt for Sub-Query Refinement.}
    \vspace{-8pt}
\label{tab:reflection_prompt}
\scriptsize

\begin{tabular}{@{}p{\linewidth}@{}}
\toprule
\textbf{Sub-Query Refinement Prompt} \\
\midrule

You are refining sub-questions for a video question. Given the current
sub-questions and newly retrieved video frames, revise the sub-questions
to better capture the information needed to answer the main question.

\smallskip
\textbf{Each refined sub-question must:}
\begin{itemize}[leftmargin=*, noitemsep, topsep=0pt]
    \item Use the retrieved visual evidence to preserve useful
    sub-questions, revise ineffective ones, or introduce missing
    information needs.

    \item Be answerable by looking at a single still frame---ask about
    visible objects, people, places, attributes, settings, or counts.
    Avoid asking about actions, sequences, order, changes over time,
    or ``what happens next''.

    \item Point to information the main question implicitly depends on,
    but does not itself ask about directly.

    \item Be different from the main question and from each other.
\end{itemize}

\smallskip
\textbf{Main question:} \texttt{\{query\}} \\
\textbf{Current sub-questions:} \texttt{\{sub\_queries\}} \\
\bottomrule
\end{tabular}
\end{minipage}

\end{table*}

\subsubsection{Tool-Based Evidence Localization}
\label{sec:method_retrieval}

As discussed in Section~\ref{sec:challenge_selection}, the Interpretation--Selection Gap limits the reliability of direct VLM relevance judgments, while inference costs increase with the candidate pool size. To address this gap, the Tool-Based Evidence Localization module delegates frame selection to an embedding model, using query--frame similarity to retrieve relevant frames from the candidate pool.

Specifically, the embedding model separately matches the original query $q$ and generated sub-queries ${q_1,\ldots,q_M}$ against each candidate frame $I_t$ through query--frame similarity:
\begin{equation}
S(q,I_t)=\cos\!\left(E_t(q),E_v(I_t)\right), \qquad
S(q_m,I_t)=\cos\!\left(E_t(q_m),E_v(I_t)\right),\quad m=1,\ldots,M,
\label{eq:similarity}
\end{equation}
where $E_t$ and $E_v$ denote the text and visual encoders of the embedding model, respectively.

The resulting similarity scores are adaptively fused into a relevance score $s_t$ for each candidate frame:
\begin{equation}
s_t=w_0S(q,I_t)+\sum_{m=1}^{M}w_mS(q_m,I_t),
\qquad
w_m=(1-w_0)\frac{\exp(-\beta H_m)}{\sum_{j=1}^{M}\exp(-\beta H_j)}.
\label{eq:fusion}
\end{equation}
Here, $w_0$ is the fixed weight for the original query, $w_m$ is the adaptive weight for sub-query $q_m$, and $H_m$ is the entropy of its query--frame similarity distribution. The parameter $\beta$ controls entropy sensitivity, with larger weights assigned to lower-entropy, more discriminative retrieval signals.

Finally, DPP-based sampling~\citep{dpp-1,dpp-2} selects $N$ relevant and visually diverse frames by greedily approximating determinant maximization over $L_{\mathcal{K}}$, the submatrix of the DPP kernel $L$ indexed by $\mathcal{K}$:
\begin{equation}
\mathcal{K}\approx\underset{|\mathcal{K}|=N}{\arg\max}\det(L_{\mathcal{K}}),
\qquad
L_{ij}=(s_i s_j)^{\alpha}\frac{1+\cos\!\left(E_v(I_i),E_v(I_j)\right)}{2}.
\label{eq:dpp}
\end{equation}
Here, the kernel $L$ combines min-max-normalized relevance scores $s_t$ with rescaled cosine similarity to capture visual redundancy. Larger $\alpha$ emphasizes relevance, while smaller $\alpha$ favors visual diversity.

\subsubsection{Reflection Loop}
\label{sec:method_reflection}
To iteratively enhance Query Interpretation and Tool-Based Evidence Localization, RACER establishes a reflection loop in which tool-selected key frames serve as visual feedback for sub-query refinement.

Specifically, in each reflection round, the first $B$ frames selected by DPP-based sampling form the feedback set $\mathcal{R}$. A hybrid overview $\mathcal{H}$ combines all feedback frames with selected uniform overview frames under the budget $N$:
\begin{equation}
\mathcal{H} = \mathcal{R}\cup \mathrm{Top}_{N-|\mathcal{R}|} \left(\mathrm{Uniform}(\mathcal{V},N)\setminus\mathcal{R};\,r_u\right),
\qquad
r_u=\max_{I_r\in\mathcal{R}}
\frac{1+\cos\!\left(E_v(I_u),E_v(I_r)\right)}{2}.
\label{eq:hybrid}
\end{equation}
After excluding frames already in $\mathcal{R}$, each remaining overview frame $I_u$ is scored by its maximum visual similarity to any feedback frame $I_r$. The operator $\mathrm{Top}_{k}(\cdot;r_u)$ retains the $k$ highest-scoring frames, removing the least similar overview frames while preserving all feedback frames. Thus, $\mathcal{H}$ combines retrieved evidence with visually aligned overview context.

Finally, given the original query $q$, hybrid overview $\mathcal{H}$, and current sub-queries$\{q_1,\ldots,q_M\}$, the lightweight Vid-LLM $\pi$ refines the retrieval targets:
\begin{equation}
\{q'_1,\ldots,q'_M\}
=
\pi\!\left(q,\mathcal{H},\{q_1,\ldots,q_M\}\right).
\label{eq:reflect}
\end{equation}
The refinement is guided by the prompt in Table~\ref{tab:reflection_prompt}, which instructs $\pi$ to revise sub-queries using retrieved evidence while preserving complementarity and visual grounding. The refined sub-queries guide the next Tool-Based Evidence Localization round, closing the reflection loop.

\newcommand{\gain}[1]{\textcolor{green!50!black}{(#1)}}

\begin{table}[t]
\caption{
Comparison with state-of-the-art frame-selection methods across three benchmarks and four Vid-LLMs: accuracy (\%) for Video-MME and LongVideoBench, and M-Avg for MLVU.
Parentheses show gains over uniform sampling (positive in green).
Average is the unweighted mean; \textbf{bold}/\underline{underline} mark best/second-best results per Vid-LLM. RACER achieves the best or tied-best performance across all benchmarks and answering models.
}
\label{tab:main}
\centering
\resizebox{\linewidth}{!}{%
\begin{tabular}{l|lccc|c}
\toprule
\textbf{Vid-LLMs} &
\textbf{Frame Selection Method} &
\textbf{Video-MME} (1--60 min) &
\textbf{MLVU} (3--120 min) &
\textbf{LongVideoBench} (1--60 min) &
\textbf{Average} \\
\midrule

\multirow{8}{*}{LLaVA-OV-7B}
& Uniform
& 58.5 (+0.0)
& 62.4 (+0.0)
& 56.6 (+0.0)
& 59.2 (+0.0) \\

& AKS~\citep{aks}
& 58.4 (-0.1)
& 66.8 \gain{+4.4}
& 59.3 \gain{+2.7}
& 61.5 \gain{+2.3} \\

& BOLT~\citep{bolt}
& 59.9 \gain{+1.4}
& 66.8 \gain{+4.4}
& 59.6 \gain{+3.0}
& 62.1 \gain{+2.9} \\

& MDP$^3$~\citep{mdp3}
& 60.5 \gain{+2.0}
& 68.3 \gain{+5.9}
& 59.0 \gain{+2.4}
& 62.6 \gain{+3.4} \\

& AdaQ~\citep{adaq}
& 59.5 \gain{+1.0}
& \underline{69.4} \gain{+7.0}
& 60.0 \gain{+3.4}
& 63.0 \gain{+3.8} \\

& GIFT~\citep{gift}
& 61.2 \gain{+2.7}
& \underline{69.4} \gain{+7.0}
& 59.6 \gain{+3.0}
& 63.4 \gain{+4.2} \\

& A.I.R.~\citep{air}
& \underline{61.4} \gain{+2.9}
& 69.3 \gain{+6.9}
& \underline{60.7} \gain{+4.1}
& \underline{63.8} \gain{+4.6} \\

& \cellcolor{gray!20}RACER (ours)
& \cellcolor{gray!20}\textbf{61.9} \gain{+3.4}
& \cellcolor{gray!20}\textbf{70.0} \gain{+7.6}
& \cellcolor{gray!20}\textbf{60.9} \gain{+4.3}
& \cellcolor{gray!20}\textbf{64.3} \gain{+5.1} \\

\midrule

\multirow{5}{*}{LLaVA-Video-7B}
& Uniform
& 62.6 (+0.0)
& 64.7 (+0.0)
& 58.0 (+0.0)
& 61.8 (+0.0) \\

& AKS~\citep{aks}
& 64.5 \gain{+1.9}
& 65.4 \gain{+0.7}
& 59.8 \gain{+1.8}
& 63.2 \gain{+1.4} \\

& BOLT~\citep{bolt}
& 64.1 \gain{+1.5}
& 66.5 \gain{+1.8}
& 59.8 \gain{+1.8}
& 63.5 \gain{+1.7} \\

& GIFT~\citep{gift}
& \textbf{65.0} \gain{+2.4}
& \underline{67.4} \gain{+2.7}
& \underline{60.2} \gain{+2.2}
& \underline{64.2} \gain{+2.4} \\

& \cellcolor{gray!20}RACER (ours)
& \cellcolor{gray!20}\textbf{65.0} \gain{+2.4}
& \cellcolor{gray!20}\textbf{72.0} \gain{+7.3}
& \cellcolor{gray!20}\textbf{61.3} \gain{+3.3}
& \cellcolor{gray!20}\textbf{66.1} \gain{+4.3} \\

\midrule

\multirow{7}{*}{Qwen2.5-VL-7B}
& Uniform
& 60.8 (+0.0)
& 59.3 (+0.0)
& 58.1 (+0.0)
& 59.4 (+0.0) \\

& AKS~\citep{aks}
& 62.8 \gain{+2.0}
& \underline{69.1} \gain{+9.8}
& 58.8 \gain{+0.7}
& 63.6 \gain{+4.2} \\

& BOLT~\citep{bolt}
& 64.1 \gain{+3.3}
& 66.3 \gain{+7.0}
& 58.6 \gain{+0.5}
& 63.0 \gain{+3.6} \\

& MDP$^3$~\citep{mdp3}
& 63.8 \gain{+3.0}
& 66.2 \gain{+6.9}
& 60.0 \gain{+1.9}
& 63.3 \gain{+3.9} \\

& GIFT~\citep{gift}
& \underline{64.4} \gain{+3.6}
& 68.2 \gain{+8.9}
& 61.3 \gain{+3.2}
& \underline{64.6} \gain{+5.2} \\

& A.I.R.~\citep{air}
& \textbf{65.0} \gain{+4.2}
& 67.5 \gain{+8.2}
& \underline{61.4} \gain{+3.3}
& \underline{64.6} \gain{+5.2} \\

& \cellcolor{gray!20}RACER (ours)
& \cellcolor{gray!20}\textbf{65.0} \gain{+4.2}
& \cellcolor{gray!20}\textbf{70.9} \gain{+11.6}
& \cellcolor{gray!20}\textbf{62.2} \gain{+4.1}
& \cellcolor{gray!20}\textbf{66.0} \gain{+6.6} \\

\midrule

\multirow{4}{*}{Qwen3-VL-8B}
& Uniform
& 64.4 (+0.0)
& 64.4 (+0.0)
& 60.4 (+0.0)
& 63.1 (+0.0) \\

& BOLT~\citep{bolt}
& 65.7 \gain{+1.3}
& 70.3 \gain{+5.9}
& 62.2 \gain{+1.8}
& 66.1 \gain{+3.0} \\

& AdaQ~\citep{adaq}
& \underline{67.7} \gain{+3.3}
& \underline{72.0} \gain{+7.6}
& \underline{63.8} \gain{+3.4}
& \underline{67.8} \gain{+4.7} \\

& \cellcolor{gray!20}RACER (ours)
& \cellcolor{gray!20}\textbf{67.9} \gain{+3.5}
& \cellcolor{gray!20}\textbf{73.5} \gain{+9.1}
& \cellcolor{gray!20}\textbf{64.0} \gain{+3.6}
& \cellcolor{gray!20}\textbf{68.5} \gain{+5.4} \\

\bottomrule
\end{tabular}}
\end{table}

\section{Experiments}
\label{sec:exps}

\subsection{Experimental Setup}
\label{sec:exp_setup}

To evaluate RACER for long video frame selection, we conduct experiments on Video-MME (w/o subtitles)~\citep{videomme}, MLVU~\citep{mlvu}, and LongVideoBench~\citep{longvideobench} using lmms-eval~\citep{lmmseval}. Following official evaluation protocols, we report accuracy for Video-MME and LongVideoBench and M-Avg for MLVU. Four representative Vid-LLMs serve as answering models: LLaVA-OneVision~\citep{llavaov}, LLaVA-Video~\citep{llavavideo}, Qwen2.5-VL~\citep{qwen25vl}, and Qwen3-VL~\citep{qwen3vl}, each with a frame budget of $N=32$.
RACER uses Qwen2.5-VL-3B for query interpretation and CLIP ViT-L/14 as the retrieval tool. Candidate frames are sampled at 2 fps for videos shorter than 120 seconds and 1 fps otherwise. Unless otherwise specified, we use $M=3$ sub-queries, a feedback budget of $B=24$, two reflection rounds, $w_0=0.6$, $\alpha=0.75$, and $\beta=2.2$. All components remain frozen, with no training required. Experiments are conducted on $8\times$A100-80GB GPUs.

\subsection{Comparison to the State-of-the-Art}
\label{sec:exp_main}

As shown in Table~\ref{tab:main}, RACER consistently improves long-video understanding across all four answering Vid-LLMs, achieving the best average performance in every setting. RACER yields average gains of $4.3$--$6.6$ over uniform sampling and consistently outperforms prior frame-selection methods across the three benchmarks.
The gains are particularly pronounced on MLVU, which contains the longest videos of the three benchmarks, up to 120 minutes. RACER improves over uniform sampling by $7.3$--$11.6$ on MLVU. These results support the effectiveness of decomposing frame selection into query interpretation and evidence localization, with each task assigned to a distinct paradigm.

\subsection{Ablation Studies}
\label{sec:exp_module}

\textbf{Module Ablation.}
To assess each component's contribution, we conduct cumulative ablations on Video-MME with LLaVA-OV-7B as the answering model. As shown in Table~\ref{tab:module}, Tool-Based Retrieval improves overall accuracy from $58.5$ to $60.6$. Adding Query Interpretation raises accuracy to $61.5$, and Reflection further improves it to $61.9$, with the full framework achieving the best performance across all three video-duration groups. Notably, despite lower direct answering accuracy than LLaVA-OV-7B ($57.8$ vs. $58.5$ under uniform sampling), Qwen2.5-VL-3B enhances the stronger model when assigned query interpretation. These results support decomposing frame selection into query interpretation and evidence localization, with task assignments aligned with model capabilities, enabling limited-capability components to enhance more capable Vid-LLMs.

\begin{table}[t]
\caption{
Module ablation of RACER on Video-MME with components added cumulatively.
Scores report accuracy (\%) by duration and overall; \textbf{bold} marks column bests. Each component improves performance, with full RACER performing best.
Despite weaker direct answering capability, Qwen2.5-VL-3B improves LLaVA-OV-7B when assigned Query Interpretation.
}
\label{tab:module}
\begin{center}
\small
\resizebox{\linewidth}{!}{%
\begin{tabular}{llcccccc}
\toprule
\multirow[c]{2}{*}[-0.6ex]{\textbf{Answering Vid-LLM}} &
\multirow[c]{2}{*}[-0.6ex]{\textbf{Frame Selection}} &
\multirow[c]{2}{*}[-0.6ex]{\textbf{Retrieval Tool}} &
\multirow[c]{2}{*}[-0.6ex]{\textbf{\shortstack{Query Interpretation Model}}} &
\multicolumn{4}{c}{\textbf{Video-MME}} \\
\cmidrule(lr){5-8}
& & & & \textbf{Short} & \textbf{Medium} & \textbf{Long} & \textbf{Overall} \\
\midrule
Qwen2.5-VL-3B
& Uniform Sampling
& N/A
& N/A
& 69.5 & 57.0 & 47.1 & 57.8 \\
\midrule
\multirow{4}{*}{LLaVA-OV-7B}
& Uniform Sampling
& N/A
& N/A
& 70.2 & 57.1 & 48.2 & 58.5 \\
& + Tool-based Retrieval
& CLIP ViT-L/14
& N/A
& 71.2 & 58.7 & 51.8 & 60.6 \\
& + Query Interpretation
& CLIP ViT-L/14
& Qwen2.5-VL-3B
& 72.3 & 60.1 & 52.1 & 61.5 \\
& \cellcolor{gray!20}+ Reflection (RACER)
& \cellcolor{gray!20}CLIP ViT-L/14
& \cellcolor{gray!20}Qwen2.5-VL-3B
& \cellcolor{gray!20}\textbf{72.6}
& \cellcolor{gray!20}\textbf{60.6}
& \cellcolor{gray!20}\textbf{52.7}
& \cellcolor{gray!20}\textbf{61.9} \\
\bottomrule
\end{tabular}}
\end{center}
\end{table}

\begin{table*}[t]
\centering

\begin{minipage}[t]{0.49\textwidth}
\captionof{table}{
Component replacement on Video-MME with LLaVA-OV-7B.
Acc.: accuracy (\%); \textbf{bold}: best per group.
RACER performs consistently across retrieval tools and Query Interpretation models.
}
\label{tab:component}
\centering
\small
\resizebox{\linewidth}{!}{%
\begin{tabular}{lcc}
\toprule
\textbf{Retrieval Tool} &
\textbf{Query Interpretation Model} &
\textbf{Acc. (\%)} \\
\midrule
CLIP ViT-L/14
& Qwen2.5-VL-3B
& 61.93 \\

CLIP ViT-L/14
& Qwen3-VL-4B
& \textbf{62.04} \\

\midrule

SigLIP-L/16
& Qwen2.5-VL-3B
& \textbf{62.26} \\

EVA02-CLIP-L/14
& Qwen2.5-VL-3B
& 61.81 \\

\bottomrule
\end{tabular}}
\end{minipage}
\hfill
\begin{minipage}[t]{0.49\textwidth}
\captionof{table}{
Cumulative efficiency ablation on Video-MME with LLaVA-OV-7B.
Acc.: accuracy (\%); \# Frames: mean distinct Vid-LLM input frames; Time: seconds.
RACER improves accuracy with limited frame analysis.
}
\label{tab:module_efficiency}
\centering
\small
\resizebox{\linewidth}{!}{%
\begin{tabular}{lccc}
\toprule
\textbf{Frame Selection} & \textbf{Acc. (\%)} & \textbf{\# Frames} & \textbf{Time (s)} \\
\midrule
Uniform Sampling
& 58.5 & 0.0 & 1.1 \\
\midrule

+ Tool-based Retrieval
& 60.6 & 0.0 & 15.1 \\

+ Query Interpretation
& 61.5 & 32.0 & 22.4 \\

\rowcolor{gray!20}
+ Reflection (full RACER)
& \textbf{61.9} & 62.9 & 35.4 \\
\bottomrule
\end{tabular}}
\end{minipage}

\end{table*}

\textbf{Component Ablation.}
To assess sensitivity to component choices, we replace the retrieval tool or Query Interpretation model while keeping the remaining framework unchanged. As shown in Table~\ref{tab:component}, replacing CLIP ViT-L/14 with SigLIP-L/16 or EVA02-CLIP-L/14 yields accuracies of $62.26$ and $61.81$, compared with $61.93$ for the default configuration. Replacing Qwen2.5-VL-3B with Qwen3-VL-4B improves accuracy to $62.04$. Comparable performance across configurations supports the applicability of RACER's task decomposition to different models for query interpretation and evidence localization.

\textbf{Efficiency Ablation.}
To analyze RACER's computational overhead, we measure accuracy, inference time, and the number of distinct frames analyzed by the lightweight Vid-LLM across all rounds as components are added cumulatively on Video-MME. As shown in Table~\ref{tab:module_efficiency}, Tool-Based Retrieval takes $15.1$ seconds without requiring Vid-LLM frame analysis. Query Decomposition processes 32 overview frames, increasing runtime to $22.4$ seconds. With Reflection, the total reaches an average of $62.9$ distinct frames and $35.4$ seconds. These results quantify the additional runtime of query interpretation and reflection, while showing that candidate-pool scoring is delegated to CLIP and Vid-LLM input remains limited to a compact set of distinct frames.

\begin{figure*}[t]
\centering

\begin{subfigure}[t]{0.98\textwidth}
    \centering
    \includegraphics[width=\textwidth]{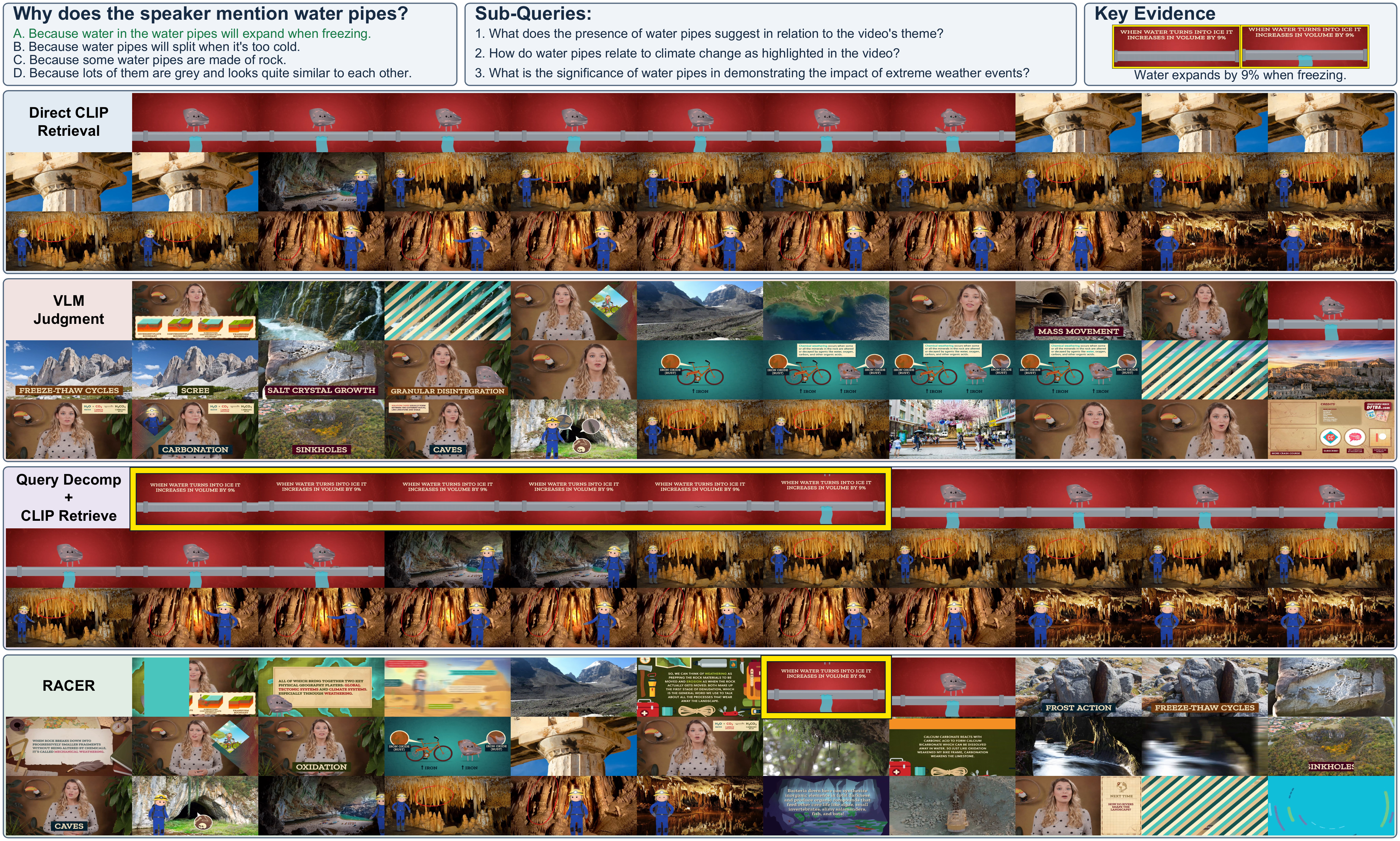}
    \caption{Query Interpretation and Tool-Based Evidence Localization.}
    \label{fig:visualization_comprehension}
\end{subfigure}

\begin{subfigure}[t]{0.98\textwidth}
    \centering
    \includegraphics[width=\textwidth]{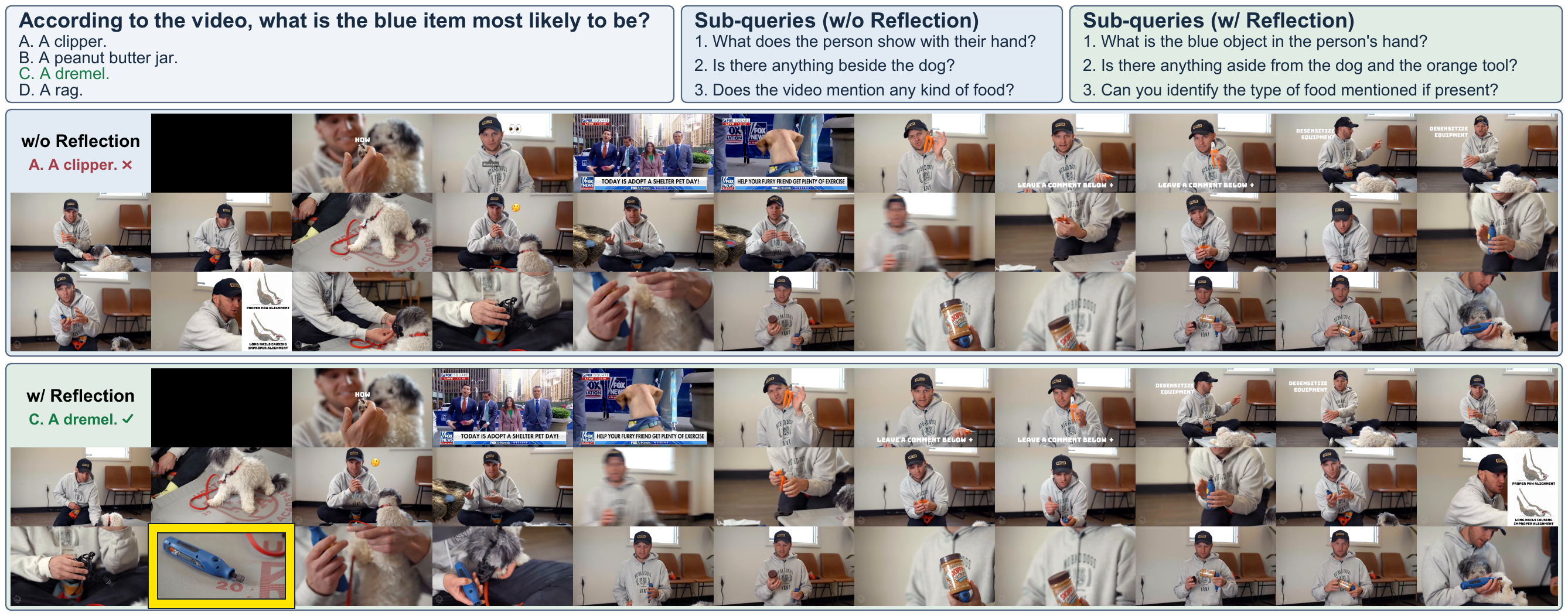}
    \caption{Reflection Loop.}
    \label{fig:visualization_reflection}
\end{subfigure}
\vspace{-8pt}
\caption{
\textbf{Qualitative visualization of RACER.}
\textbf{(a)} Query interpretation guides tool-based evidence localization to recover the explanation that water expands upon freezing, missed by direct CLIP retrieval and VLM judgment. Full RACER retains this evidence with less redundancy.
\textbf{(b)} Reflection refines sub-queries to target the blue object, recovering a previously missed close-up and correcting the answer from \emph{a clipper} to \emph{a dremel}.
}
\label{fig:visualization}
\vspace{-10pt}
\end{figure*}

\subsection{Qualitative Results}
Figure~\ref{fig:visualization_comprehension} illustrates the benefits of decomposing frame selection into query interpretation and evidence localization. Direct CLIP retrieval selects repetitive frames, whereas VLM judgment captures diverse content relevant to implicit query requirements, such as geological processes. Both, however, miss the decisive evidence. Query Interpretation enables CLIP to retrieve the explanation that water expands upon freezing, and full RACER retains this evidence with less redundancy. Beyond this division of responsibilities, reflection further improves frame selection, as shown in Figure~\ref{fig:visualization_reflection}. Broad sub-queries initially miss the decisive close-up, leading to the incorrect answer, \emph{a clipper}. Reflection refines the target to the blue object in the person's hand, distinguishing it from the dog and orange tool. The updated sub-queries recover a close-up of the tool and its sanding attachment, supporting the correct answer, \emph{a dremel}.

\section{Conclusion}
\label{sec:conclusion}

This paper investigates long-video frame selection from a task-decomposition perspective, identifying the Query Comprehension Gap in similarity-based methods and the Interpretation--Selection Gap in judgment-based methods.
To address these gaps, RACER decomposes frame selection into query interpretation and evidence localization within a training-free reflective agentic framework. A lightweight Vid-LLM makes implicit evidence requirements explicit through query decomposition, while an embedding-based retrieval tool selects relevant frames. Reflection on retrieved evidence refines the sub-queries, iteratively improving frame selection.
Experiments across three benchmarks and four answering Vid-LLMs demonstrate consistent gains in long-video understanding. These findings support assigning query interpretation and evidence localization to models with suitable capabilities, enabling limited-capability components to enhance more capable Vid-LLMs.

Future work will explore open-ended video understanding and more efficient reflection strategies to reduce computational overhead.

\section*{AI Use Statement}
Generative AI assisted with two aspects of this work. First, AI assistants edited author-written drafts for grammar, clarity, and readability. Second, AI coding assistants supported method implementation and debugging based on author-defined ideas and designs. The authors are responsible for reviewing AI-assisted edits, testing code, and verifying outputs, and take full responsibility for the final text, claims, and artifacts.

\section*{Ethics Statement}
RACER is a training-free framework that decomposes long-video frame selection into query interpretation and evidence localization. Experiments use publicly available benchmarks and pretrained models without additional training. RACER may inherit model biases and inaccuracies, and omitted evidence may produce incomplete or incorrect answers. Responsible use should respect privacy and dataset and model usage conditions. Human oversight remains important in safety-sensitive applications, as benchmark improvements do not guarantee reliable real-world performance.

\section*{Reproducibility Statement}
RACER uses publicly available datasets and pretrained models and requires no additional training. The paper specifies model configurations, evaluation protocols, frame budgets, hyperparameters, and the iterative frame-selection procedure, including prompts for query interpretation and sub-query refinement. Ablation studies examine module contributions, component choices, and computational overhead. We will release the code, experimental configurations, and per-question outputs to support reproduction and further research.

\bibliography{references}
\bibliographystyle{racer}

\newpage
\appendix

\section{Comparison across Question Types}
\label{app:additional_results}

To assess RACER's effectiveness across question types, Table~\ref{tab:task_breakdown} compares uniform sampling, direct CLIP selection, VLM judgment-based selection, and RACER across six Video-MME categories using LLaVA-OV-7B as the answering model. Category-level accuracy is reported over all 2700 questions. RACER achieves the highest accuracy across all six categories, demonstrating consistent benefits across diverse query requirements.

\begin{table*}[t]
\caption{
Accuracy (\%) across six Video-MME question categories with LLaVA-OV-7B.
}
\label{tab:task_breakdown}
\centering
\small
\setlength{\tabcolsep}{5pt}
\resizebox{\textwidth}{!}{%
\begin{tabular}{lcccccc}
\toprule
Method
& Perception
& Recognition
& Reasoning
& \shortstack{Information Synopsis}
& \shortstack{OCR Problems}
& \shortstack{Counting Problem} \\
\midrule
Uniform
& 68.9 & 60.8 & 53.2 & 74.3 & 66.2 & 36.6 \\
CLIP Retrieve
& 69.2 & 59.5 & 53.9 & 66.9 & 66.2 & 37.3 \\
VLM Judgment
& 70.7 & 62.1 & 55.8 & 75.2 & 66.2 & 36.2 \\
\rowcolor{gray!15}
RACER (ours)
& \textbf{71.9}
& \textbf{63.4}
& \textbf{57.8}
& \textbf{75.8}
& \textbf{67.6}
& \textbf{41.0} \\
\bottomrule
\end{tabular}}
\end{table*}

\begin{table*}[t]
\caption{
Efficiency comparison on Video-MME across video durations.
Time is measured in seconds, and \#Frames denotes the average number of unique frames analyzed by the VLM.
}
\label{tab:duration_efficiency}
\centering
\small
\setlength{\tabcolsep}{3pt}
\resizebox{\linewidth}{!}{%
\begin{tabular}{lcccccccccccc}
\toprule
\multirow{2}{*}{Method}
& \multicolumn{3}{c}{Short (1--3 min)}
& \multicolumn{3}{c}{Medium (3--30 min)}
& \multicolumn{3}{c}{Long (30--60 min)}
& \multicolumn{3}{c}{Overall} \\
\cmidrule(lr){2-4}
\cmidrule(lr){5-7}
\cmidrule(lr){8-10}
\cmidrule(lr){11-13}
& Acc. & Time (s) & \#Frames
& Acc. & Time (s) & \#Frames
& Acc. & Time (s) & \#Frames
& Acc. & Time (s) & \#Frames \\
\midrule

Uniform
& 70.2 & 1.1 & 0
& 57.1 & 1.1 & 0
& 48.2 & 1.1 & 0
& 58.5 & 1.1 & 0 \\

CLIP Retrieval
& 68.8 & 3.5 & 0
& 58.6 & 7.5 & 0
& 45.8 & 32.8 & 0
& 57.7 & 14.6 & 0 \\

VLM judgment
& 70.7 & 22.8 & 157.7
& 58.0 & 58.4 & 516.3
& 51.7 & 286.2 & 2465.9
& 60.1 & 122.4 & 1046.6 \\

\rowcolor{gray!20}
RACER
& \textbf{72.6} & 22.4 & 57.6
& \textbf{60.6} & 29.0 & 64.0
& \textbf{52.7} & 54.9 & 67.1
& \textbf{61.9} & 35.4 & 62.9 \\

\bottomrule
\end{tabular}%
}
\end{table*}

\section{Efficiency across Video Durations}
Table~\ref{tab:duration_efficiency} compares frame-selection methods on Video-MME across three video-duration groups. RACER achieves the highest accuracy in every group while scaling more efficiently than VLM judgment-based selection. As videos increase from short to long, VLM judgment-based selection analyzes $157.7$ to $2465.9$ distinct frames, with inference time rising from $22.8$ to $286.2$ seconds. RACER analyzes only $57.6$--$67.1$ distinct frames, with runtime increasing from $22.4$ to $54.9$ seconds. On long videos, RACER achieves a $5.2\times$ speedup while improving accuracy from $51.7$ to $52.7$. Overall, RACER improves accuracy from $60.1$ to $61.9$ with a $3.5\times$ speedup. These results demonstrate the scalability benefits of delegating candidate-pool scoring to a retrieval tool while limiting Vid-LLM analysis to a compact set of frames.

\section{Additional Hyperparameter Ablation}
\label{app:hp}

Table~\ref{tab:hp_sensitivity} examines the DPP relevance parameter $\alpha$, entropy sensitivity $\beta$, original-query weight $w_0$, and lightweight Vid-LLM generation temperature $\tau$ on Video-MME with LLaVA-OV-7B. Each sweep varies one parameter, with default settings marked by $\dagger$ and the best observed result in bold.

\subsection{Effect of \texorpdfstring{$\alpha$}{alpha} in DPP Sampling}
\label{app:hp_alpha}

The parameter $\alpha$ balances relevance and visual diversity in DPP sampling. Accuracy increases from $60.41\%$ at $\alpha=0.55$ to a peak of $61.93\%$ at $\alpha=0.75$, while larger values yield no further improvement. The results favor balancing relevance and diversity over placing excessive emphasis on relevance.

\begin{table*}[t]
\caption{
Hyperparameter ablation on Video-MME with LLaVA-OV-7B.
Each column pair reports a separate parameter sweep.
Acc. denotes accuracy (\%); $\dagger$ marks the default setting; bold indicates the best observed result.
}
\label{tab:hp_sensitivity}
\centering
\small
\setlength{\tabcolsep}{9pt}
\begin{tabular}{cccccccc}
\toprule
\multicolumn{2}{c}{DPP Relevance} &
\multicolumn{2}{c}{Entropy Sensitivity} &
\multicolumn{2}{c}{Original-Query Weight} &
\multicolumn{2}{c}{Generation Temperature} \\
\cmidrule(lr){1-2}\cmidrule(lr){3-4}\cmidrule(lr){5-6}\cmidrule(lr){7-8}
$\alpha$ & Acc. & $\beta$ & Acc. & $w_0$ & Acc. & $\tau$ & Acc. \\
\midrule
0.55 & 60.41 & 2.0 & \textbf{62.11} & 0.4 & 61.37 & 0.6 & 60.30 \\
0.65 & 60.63 & 2.1 & 61.96 & 0.5 & 61.22 & 0.7 & 61.00 \\
$0.75^{\dagger}$ & \textbf{61.93} & $2.2^{\dagger}$ & 61.93 & $0.6^{\dagger}$ & \textbf{61.93} & $0.8^{\dagger}$ & \textbf{61.93} \\
0.85 & 60.89 & 2.3 & 61.89 & 0.7 & 61.22 & 0.9 & 60.59 \\
0.95 & 61.52 & 2.4 & 61.93 & 0.8 & 60.48 & 1.0 & 60.89 \\
\bottomrule
\end{tabular}
\end{table*}

\begin{table}[t]
\caption{
Ablations of sub-query aggregation, reflection depth, and reflection input on Video-MME with LLaVA-OV-7B. Acc. denotes accuracy (\%). $\dagger$ marks the default setting, and bold indicates the best result within each group. Each reflection step refines the sub-queries after retrieval; the default two steps yield three retrieval rounds, including the initial round.
}
\label{tab:design_ablation}
\centering
\small
\setlength{\tabcolsep}{6pt}
\begin{tabular}{llc}
\toprule
Design & Configuration & Acc. (\%) \\
\midrule
Sub-query aggregation & Uniform weighting & 61.67 \\
& Entropy-based weighting$^{\dagger}$ & \textbf{61.93} \\
\midrule
Reflection steps & 0 & 61.52 \\
& 1 & 61.89 \\
& $2^{\dagger}$ & 61.93 \\
& 3 & \textbf{61.96} \\
& 4 & \textbf{61.96} \\
\midrule
Reflection input & Without reflection & 61.52 \\
& Uniform overview & 61.56 \\
& Hybrid overview (temporal-distance pruning) & 61.70 \\
& Hybrid overview (visual-similarity pruning)$^{\dagger}$ & \textbf{61.93} \\
\bottomrule
\end{tabular}
\end{table}

\subsection{Effect of \texorpdfstring{$\beta$}{beta} in Entropy-Based Fusion}
\label{app:hp_beta}

The parameter $\beta$ controls the influence of entropy on sub-query weights, with larger values favoring lower-entropy retrieval signals. Accuracy varies only from $61.89\%$ to $62.11\%$ over $\beta\in[2.0,2.4]$. Although $\beta=2.0$ yields the highest accuracy, the default $\beta=2.2$ achieves a comparable $61.93\%$, indicating limited sensitivity within the tested range.

\subsection{Effect of the Original-Query Weight \texorpdfstring{$w_0$}{w0}}
\label{app:hp_w0}

The weight $w_0$ balances the original-query score against the combined sub-query scores. The default $w_0=0.6$ achieves the highest accuracy of $61.93\%$, while both lower and higher weights reduce performance. In particular, accuracy drops to $60.48\%$ at $w_0=0.8$, supporting the use of sub-queries alongside the original query.

\subsection{Effect of the Vid-LLM Generation Temperature}
\label{app:hp_temperature}

The temperature $\tau$ controls sampling during sub-query generation and refinement. The default $\tau=0.8$ achieves the highest accuracy of $61.93\%$, compared with $60.30\%$ at $\tau=0.6$ and $60.89\%$ at $\tau=1.0$. The non-monotonic trend indicates that neither lower nor higher generation temperature consistently improves frame selection.

\section{Additional Framework Design Ablations}
\label{app:design_ablation}

Table~\ref{tab:design_ablation} evaluates RACER's design choices for sub-query aggregation, reflection depth, and reflection input construction on Video-MME with LLaVA-OV-7B.

\subsection{Effect of Sub-Query Aggregation}
\label{app:aggregation}

Entropy-based weighting improves accuracy from $61.67\%$ to $61.93\%$ over uniform weighting, supporting adaptive sub-query weighting based on retrieval-score distributions.

\subsection{Effect of the Number of Reflection Steps}
\label{app:reflection_steps}

Accuracy improves from $61.52\%$ without reflection to $61.89\%$ after one step and $61.93\%$ after two, reaching $61.96\%$ with three or four steps. These diminishing returns support the default two-step configuration, which captures most of the gain and requires three retrieval rounds, including the initial round.

\subsection{Effect of the Reflection Input}
\label{app:reflection_input}

Reflection with the uniform overview achieves $61.56\%$ accuracy, compared with $61.52\%$ without reflection. Hybrid overviews incorporating retrieved feedback improve accuracy to $61.70\%$ with temporal-distance pruning and $61.93\%$ with visual-similarity pruning. The default strategy retains all feedback frames and removes the least visually similar overview frames, supporting the use of retrieved evidence and visually aligned context for refinement.

\section{Additional Discussion of Related Work}
\label{app:related_work_discussion}
We further distinguish RACER from existing frame-selection methods through task decomposition. Similarity-based methods refine sampling, temporal coverage, and diversity using query--frame similarity~\citep{bolt,qframe,aks,mdp3}, whereas judgment-based methods rely on VLM/LLM decisions to assess relevance or select frames and temporal regions~\citep{videotree,llovi,videoagent,air}. Some methods combine these paradigms but retain VLM/LLM control over frame selection. VideoAgent~\citep{videoagent} combines query analysis with CLIP retrieval, with the LLM determining search regions, stopping retrieval, and adding or removing frames based on its judgments. A.I.R.~\citep{air} applies VLM relevance judgments to candidates prefiltered by CLIP. RACER instead assigns query interpretation and refinement exclusively to a lightweight Vid-LLM and delegates evidence localization entirely to an embedding-based retrieval tool. The Vid-LLM generates sub-queries without scoring frames, selecting indices, or specifying temporal regions, while the tool scores candidates and selects frames under preset budgets and reflection depth. This separation persists throughout reflection, with retrieved frames informing only sub-query refinement. RACER thus addresses the Query Comprehension Gap by making implicit evidence requirements explicit and the Interpretation--Selection Gap by delegating frame selection to the retrieval tool.

\end{document}